\def\cvprPaperID{4521}
\def\confName{CVPR}
\def\confYear{2023}

\def\authorBlock{
    $\text{Seonghoon Yu}^{1}$ \qquad
    $\text{Paul Hongsuck Seo}^{2}$ \qquad
    $\text{Jeany Son}^{1}$ \\
    $^{1}\text{AI Graduate School, GIST}$ \qquad
    $^{2}\text{Google Research}$ \\
    {\tt\small seonghoon@gm.gist.ac.kr} \qquad
    {\tt\small phseo@google.com} \qquad
    {\tt\small jeany@gist.ac.kr}
}

\newif\ifreview 
\newif\ifarxiv 
\newif\ifcamera \newcommand{\cameraready}{\cameratrue}
\newif\ifrebuttal 

\cameraready 

\pdfoutput=1
\documentclass[10pt,twocolumn,letterpaper]{article}
\usepackage{subcaption}
\usepackage{xcolor}
\usepackage{multirow}
\usepackage{mathrsfs}
\usepackage{pifont}

\usepackage[normalem]{ulem}
\usepackage[export]{adjustbox}
\usepackage[accsupp]{axessibility}

\newcommand{\ori}[1]{\textcolor{gray}{#1}}

\ifreview \usepackage[review]{cvpr} \fi
\ifarxiv \usepackage[pagenumbers]{cvpr} \fi
\ifrebuttal \usepackage[rebuttal]{cvpr} \fi
\ifcamera \usepackage{cvpr} \fi

\usepackage{graphicx}
\usepackage{amsmath}
\usepackage{amssymb}
\usepackage{booktabs}

\usepackage{times}
\usepackage{microtype}
\usepackage{epsfig}
\usepackage{caption}
\usepackage{float}
\usepackage{placeins}
\usepackage{color, colortbl}
\usepackage{stfloats}
\usepackage{enumitem}
\usepackage{tabularx}
\usepackage{xstring}
\usepackage{multirow, makecell}
\usepackage{xspace}
\usepackage{url}
\usepackage{subcaption}
\usepackage{xcolor}
\usepackage[hang,flushmargin]{footmisc}
\usepackage{mathtools}
\usepackage{arydshln}

\ifcamera \usepackage[accsupp]{axessibility} \fi

% ...

\ifarxiv  \fi

\newcommand{\R}[1]{{%
    \textbf{%
        \ifstrequal{#1}{1}{\textcolor{red}{R#1}}{%
        \ifstrequal{#1}{2}{\textcolor{blue}{R#1}}{%
        \ifstrequal{#1}{3}{\textcolor{magenta}{R#1}}{%
        \ifstrequal{#1}{4}{\textcolor{teal}{R#1}}{%
                           \textcolor{cyan}{R#1}%
        }}}}%
    }%
}}

\usepackage{xr-hyper}

\makeatletter
\newcommand*{\addFileDependency}[1]{
  \typeout{(#1)}
  \@addtofilelist{#1}
  \IfFileExists{#1}{}{\typeout{No file #1.}}
}

\makeatother

\usepackage[pagebackref,breaklinks,colorlinks]{hyperref}
\usepackage[capitalize]{cleveref}
\crefname{section}{Sec.}{Secs.}
\crefname{table}{Table}{Tables}
\crefname{figure}{Fig.}{Figs.}

\begin{document}
\title{Zero-shot Referring Image Segmentation with Global-Local Context Features}
\author{\authorBlock}
\maketitle

\begin{abstract}
Referring image segmentation (RIS) aims to find a segmentation mask given a referring expression grounded to a region of the input image.
Collecting labelled datasets for this task, however, is notoriously costly and labor-intensive.
To overcome this issue, we propose a simple yet effective zero-shot referring image segmentation method by leveraging the pre-trained cross-modal knowledge from CLIP.
In order to obtain segmentation masks grounded to the input text, we propose a mask-guided visual encoder that captures global and local contextual information of an input image.
By utilizing instance masks obtained from off-the-shelf mask proposal techniques, our method is able to segment fine-detailed instance-level groundings. 
We also introduce a global-local text encoder where 
the global feature captures complex sentence-level semantics of the entire input expression while the local feature focuses on the target noun phrase extracted by a dependency parser.
In our experiments, the proposed method outperforms several zero-shot baselines of the task and even the weakly supervised referring expression segmentation method with substantial margins.
Our code is available at \textcolor{magenta}{https://github.com/Seonghoon-Yu/Zero-shot-RIS}.
\end{abstract}
\section{Introduction}
\label{sec:intro}

Recent advances of deep learning has revolutionised computer vision and natural language processing, and addressed various tasks in the field of vision-and-language~\cite{clip, align, vilbert, lxmert, uniter, oscar, unimo}.
A key element in the recent success of the multi-modal models such as CLIP~\cite{clip} is the contrastive image-text pre-training on a large set of image and text pairs. 
It has shown a remarkable zero-shot transferability on a wide range of tasks, such as object detection~\cite{OVOD_ViLD,OVOD_prompt,OVOD_promprdet}, semantic segmentation~\cite{MaskCLIP,ZSSS_1,ZSSS_2,ZSSS_3}, image captioning~\cite{clipcap}, visual question answering (VQA)~\cite{clip+vqa} and so on.

Despite its good transferability of pre-trained multi-modal models, it is not straightforward to handle dense prediction tasks such as object detection and image segmentation.
A pixel-level dense prediction task is challenging since there is a substantial gap between the image-level contrastive pre-training task and the pixel-level downstream task such as semantic segmentation. 
There have been several attempts to reduce gap between two tasks~\cite{cris, denseclip, MaskCLIP}, but these works aim to fine-tune the model consequently requiring task-specific dense annotations, which is notoriously labor-intensive and costly. 


\begin{figure}
    \vspace{0.1cm}
    \centering
    \includegraphics[width=1\linewidth]{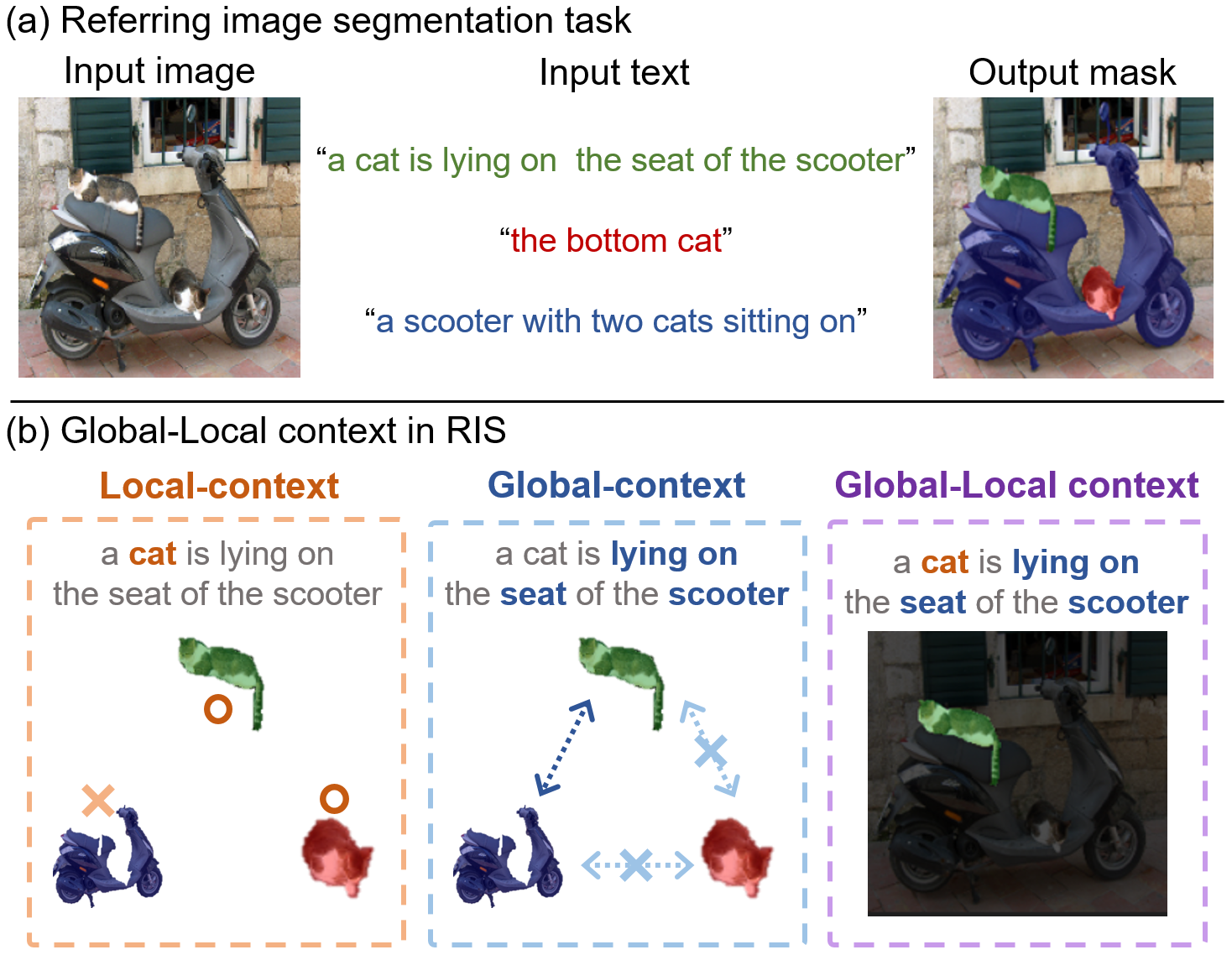}
    \caption{Illustrations of the task of referring image segmentation and motivations of global-local context features. To find the grounded mask given an expression, we need to understand the relations between the objects as well as their semantics.}
    \label{fig:intro}
\end{figure}

Referring image segmentation is a task to find the specific region in an image given a natural language text describing the region, and it is well-known as one of challenging vision-and-language tasks.
Collecting annotations for this task is even more challenging as the task requires to collect precise referring expression of the target region as well as its dense mask annotation.
Recently, a weakly-supervised referring image segmentation method~\cite{tseg} is proposed to overcome this issue.
However, it still requires high-level text expression annotations pairing with images for the target datasets and the performance of the method is far from that of the supervised methods.
To tackle this issue, in this paper, we focus on zero-shot transferring from the pre-trained knowledge of CLIP to the task of referring image segmentation.

Moreover, this task is challenging because it requires high-level understanding of language and comprehensive understanding of an image, as well as a dense instance-level prediction.
There have been several works for zero-shot semantic segmentation~\cite{MaskCLIP,ZSSS_1, ZSSS_2, ZSSS_3}, but they cannot be directly extended to the zero-shot referring image segmentation task because it has different characteristics.
Specifically, the semantic segmentation task does not need to distinguish instances, but the referring image segmentation task should be able to predict an instance-level segmentation mask. 
In addition, among multiple instances of the same class, only one instance described by the expression must be selected.
For example, in Figure~\ref{fig:intro}, there are two cats in the input image.
If the input text is given by \textit{``a cat is lying on the seat of the scooter"}, the cat with the green mask is the proper output.
To find this correct mask, we need to understand the relation between the objects (\ie \textit{``lying on the seat"}) as well as their semantics (\ie \textit{``cat", ``scooter"}).



In this paper, we propose a new baseline of zero-shot referring image segmentation task using a pre-trained model from CLIP, where global and local contexts of an image and an expression are handled in a consistent way. 
In order to localize an object mask region in an image given a textual referring expression, we propose a mask-guided visual encoder that captures global and local context information of an image given a mask.
We also present a global-local textual encoder where the local-context is captured by a target noun phrase and the global context is captured by a whole sentence of the expressions.
By combining features in two different context levels, our method is able to understand a comprehensive knowledge as well as a specific trait of the target object.
Note that, although our method does not require any additional training on CLIP model, it outperforms all baselines and the weakly supervised referring image segmentation method with a big margin.




Our main contributions can be summarised as follows:
\begin{itemize}
    \item  We propose a new task of zero-shot referring image segmentation based on CLIP without any additional training. To the best of our knowledge, this is the first work to study the zero-shot referring image segmentation task.
    
    \item We present a visual encoder and a textual encoder that integrates global and local contexts of images and sentences, respectively. Although the modalities of two encoders are different, our visual and textual features are dealt in a consistent way.
    
    \item The proposed global-local context features take full advantage of CLIP to capture the target object semantics as well as the relations between the objects in both visual and textual modalities.

    \item Our method consistently shows outstanding results compared to several baseline methods, and also outperforms the weakly supervised referring image segmentation method with substantial margins.
\end{itemize}
\section{Related Work}
\label{sec:related}

\begin{figure*}
    \centering
    \includegraphics[width=1\linewidth]{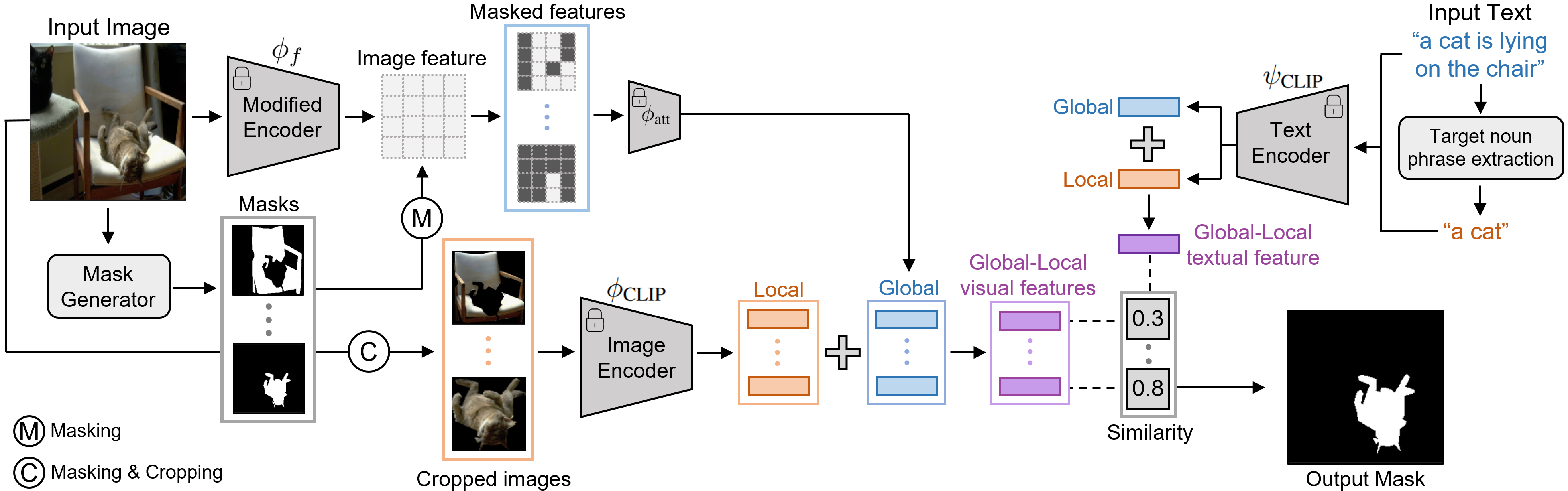}
    \caption{Overall framework of our global-local CLIP. Given an image and an expression as inputs, we extract global-local context visual features using mask proposals, and also we extract a global-local context textual feature. After computing the cosine similarity scores between all global-local context visual features and a global-local context textual feature, we choose the mask with the highest score. }
    \label{fig:framework}
\end{figure*}


\paragraph{Zero-shot Transfer.}

Classical zero-shot learning aims to predict unseen classes that have not seen before by transferring the knowledge trained on the seen classes.
Early works~\cite{contrastive, transzero, ips} leverage the pre-trained word embedding~\cite{word2vec, bert} of class names or attributes and perform zero-shot prediction via mapping between visual representations of images and this word embedding. 
Recently, CLIP~\cite{clip} and ALIGN~\cite{align} shed a new light on the zero-shot learning via large-scale image-text pre-training.
They show the successive results on various downstream tasks via zero-shot knowledge transfer, such as image captioning~\cite{clipcap}, video action localization~\cite{actionclip}, image-text retrieval~\cite{clip_retrieval} and so on.
Contrary to classical zero-shot learning, zero-shot transfer has an advantage of avoiding fine-tuning the pre-trained model on the task-specific dataset, where collecting datasets is time-consuming.
There have been several works that apply CLIP encoders directly with tiny architectural modification without additional training for semantic segmentation~\cite{MaskCLIP}, referring expression comprehension~\cite{reclip}, phrase localization~\cite{adaptingclip} and object localization~\cite{clipcam}.
Our work is also lying on the line of this research field.

\paragraph{Zero-shot Dense Prediction Tasks.}
Very recently, with the success of pre-training models using large-scale image-text pairs, there have been several attempts to deal with dense prediction tasks with CLIP, \eg object detection~\cite{OVOD_ViLD, OVOD_prompt, OVOD_promprdet, kuo2022f, lin2022learning, rasheedbridging}, semantic segmentation~\cite{denseclip, MaskCLIP, ZSSS_1, liang2022open, zhou2022zegclip, pandey2023language, kim2023zegot, luo2022segclip, xu2023side} and so on.
These dense prediction tasks, however, are challenging since CLIP learns image-level features not pixel-level fine-grained features.
In order to handle this issue, ViLD~\cite{OVOD_ViLD} introduces a method which crop the image to contain only the bounding box region, and then extract the visual features of cropped regions using CLIP to classify the unseen objects. 
This approach is applied in a wide range of dense prediction tasks which are demanded the zero-shot transfer ability of CLIP~\cite{ZSSS_1, ZSSS_2, ZSSS_3, OVOD_promprdet, OVOD_prompt, reclip}.
While this method only considers the cropped area, there are several methods~\cite{adaptingclip,MaskCLIP} to consider the global context in the image, not only just the cropped region.
Adapting CLIP~\cite{adaptingclip} proposed the phrase localization method by modifying CLIP to generate high-resolution spatial feature maps using superpixels.
MaskCLIP~\cite{MaskCLIP} modifies the image encoder of CLIP by transforming the value embedding layer and the last linear layer into two 1$\times$1 convolutional layers to handle pixel-level predictions.
In this work, we focus on extracting both global and local context visual features with CLIP.

\paragraph{Referring Image Segmentation.}
Referring image segmentation aims to segment a target object in an image given a natural linguistic expression introduced by \cite{early_3}. 
There have been several fully-supervised methods for this task, where images and expressions are used as an input, and the target mask is given for training~\cite{lavt, locate, coupalign, alignformer, polyformer, pcan}.
Most of works~\cite{lavt, restr, vlt, efn, cmsn} focuses on how to fuse those two features in different modalities extracted from independent encoders. 
Early works~\cite{early_1, early_2} extract multi-modal features by simply concatenating visual and textual features and feed them into the segmentation networks~\cite{fcn} to predict dense segmentation masks.
There have been two branches of works fusing cross-modal features; an attention based encoder fusion~\cite{efn, lavt, towards} and a cross-modal decoder fusion based on a Transformer decoder~\cite{vlt, cmsn, cris}.
Recently, a CLIP-based approach, which learns separated image and text transformer using a contrastive pre-training, has been proposed~\cite{cris}.
Those fully supervised referring image segmentation methods show good performances in general, but they require dense annotations for target masks and comprehensive expressions describing the target object.
To address this problem, TSEG~\cite{tseg} proposed a weakly-supervised referring image segmentation method which learns the segmentation model using text-based image-level supervisions.
However, this method still requires high-level referring expression annotations with images for specific datasets. 
Therefore, we propose a new baseline for zero-shot referring image segmentation without any training or supervisions.



 
\section{Method}

\label{sec:method}
In this section, we present the proposed method for zero-shot referring image segmentation in detail.
We first show an overall framework of the proposed method (\ref{subsec:overall}), and then discuss the detailed methods for extracting visual features (\ref{subsec:visfeat}) and textual features (\ref{subsec:textfeat}) to encode global and local contextual information.

\subsection{Overall Framework}
\label{subsec:overall}
To solve the task of referring image segmentation, which aims to predict {the target region grounded to the text description}, it is essential {to learn image and text representations in a shared embedding space.}
To this end, we {adopt CLIP to leverage the pre-trained cross-modal features for images and natural language.}

\begin{figure}[t]
    \centering
    \includegraphics[width=0.9\linewidth]{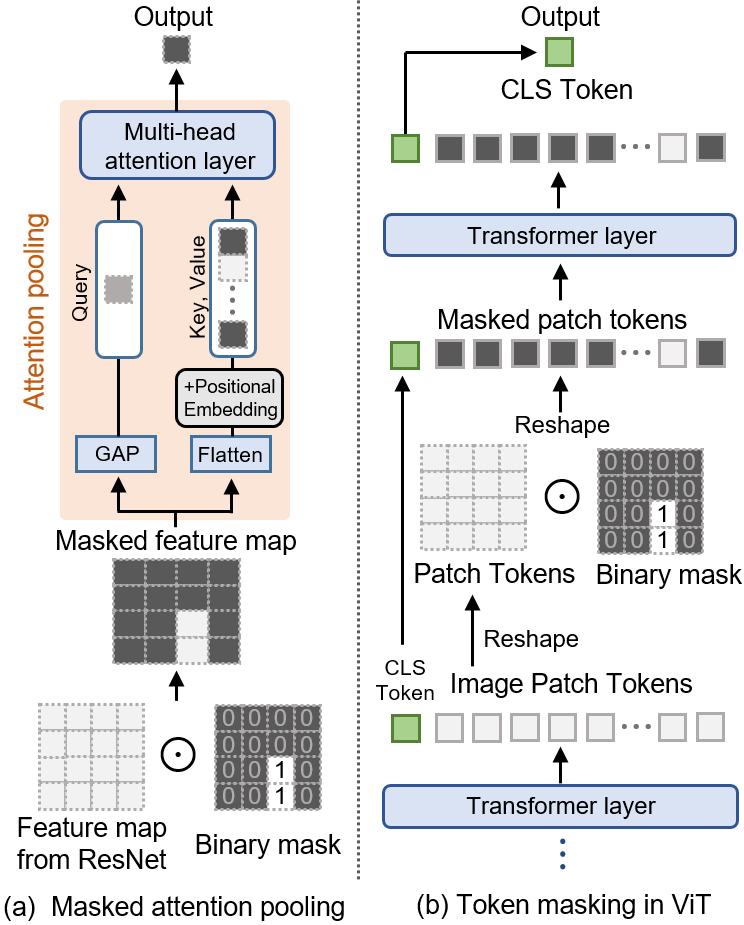}
    \caption{Detailed illustration of our mask-guided global-context visual encoders in ResNet and ViT architectures: (a) Masked attention pooling in ResNet, (b) Token masking in ViT.}
    \label{fig:global}
\end{figure}

Our framework consists of two parts as shown in Fig~\ref{fig:framework}: (1) global-local visual encoder for visual representation, and (2) global-local natural language encoder for referring expression representation.
Given a set of mask proposals generated by an unsupervised mask generator~\cite{Freesolo, cutler}, we first extract two visual features in global-context and local-context levels for each mask proposal, and then combine them into a single visual feature. 
Our global-context visual features can comprehensively represent the masked area as well as the surrounding region, while the local-context visual features can capture the representation of the specific masked region.
This acts key roles in the referring image segmentation task because we need to focus a small specific target region using a comprehensive expression of the target.
At the same time, given a sentence of expressing the target, our textual representation is extracted by the CLIP text encoder. 
In order to understand a holistic expression of the target as well as to focus on the target object itself, we first extract a key noun phrase from a sentence using a dependency parsing provided by spaCy~\cite{spacy}, and then combine a global sentence feature and a local target noun phrase feature.
Note that, our visual and text encoders are designed to handle both global-context and local-context information in a consistent way. 

Since our method is built on CLIP where the visual and textual features are embedded in the common embedding space, we can formulate the objective of our zero-shot image referring segmentation task as follows.
Given inputs of an image $I$ and a referring expression $T$, our method finds the mask that has the maximum similarity between its visual feature and the given textual feature among all mask proposals:
\begin{align}
    \hat{m} = \arg\max_{m\in M(I)}  \text{sim} (\mathbf{t}, \mathbf{f}_m),
    \label{eq:sim}
\end{align}
where $\text{sim}(\cdot,\cdot)$ is a cosine similarity, $\mathbf{t}$ is the proposed global-local textual feature for a referring expression $T$, $\mathbf{f}$ is the mask-guided global-local visual feature, and $M(I)$ is a mask proposal set for a given image $I$.

 

 
\subsection{Mask-guided Global-local Visual Features}
\label{subsec:visfeat}

To segment the target region related to the referring expression, it is essential to understand a global relationship between multiple objects in the image as well as local semantic information of the target. 
In this section, we demonstrate how to extract global and local-context features using CLIP, and how to fuse them.

Since CLIP is designed to learn image-level representation, it is not well-suited for a pixel-level dense prediction such as an image segmentation. 
To overcome the limitation of using CLIP, we decompose the task into two sub-tasks: mask proposal generation and masked image-text matching. 
In order to generate mask proposals, we use the off-the-shelf mask extractor~\cite{Freesolo} which is the unsupervised instance-level mask generation model.
By using mask proposals explicitly, our method can handle fine-detailed instance-level segmentation masks with CLIP.

\paragraph{Global-context Visual Features.}

For each mask proposals, we first extract global-context visual features using the CLIP pre-trained model.
The original visual features from CLIP, however, is designed to generate one single feature vector to describe the whole image.
To tackle this issue, we modify a visual encoder from CLIP to extract features that contain information from not only the masked region but also surrounding regions to understand relationships between multiple objects.

In this paper, we use two different architectures for the visual encoder as in CLIP: ResNet~\cite{resnet} and Vision Transformer (ViT)~\cite{vit}.
For the visual encoder with the ResNet architecture, we denote a visual feature extractor without a pooling layer as $\phi_\text{f}$ and its attention pooling layer as  $\phi_{\text{att}}$.
Then the visual feature, $\mathbf{f}$, using the visual encoder of CLIP, $\phi_\text{CLIP}$, can be expressed as follows:
\begin{align}
    \mathbf{f}=\phi_\text{CLIP}(I)=\phi_\text{att} (\phi_\text{f} (I)),
    \label{eq:clipvisfeat}
\end{align}
where $I$ is a given image.
Similarly, since ViT has multiple multi-head attention layers, we divide this visual encoder into two parts: last $k$ layers and the rest.
We denote the former one by $\phi_{\text{att}}$, and the later one by $\phi_{\text{f}}$ for ViT architectures based on CLIP.

 
 
 
 

Then given an image $I$ and a mask $m$, our global-context visual feature is defined as follows:
\begin{align}
    \mathbf{f}_m^G=\phi_{\text{att}}(\phi_f(I)\odot \bar{m}),
    \label{eq:globalvisfeat}
\end{align}
where $\bar{m}$ is the resized mask scaled to the size of the feature map, and $\odot$ is a Hadamard product operation.
We illustrate more details of this masking strategy for each architecture of CLIP in Section~\ref{subsec:masking} and Figure~\ref{fig:global}. 

We refer to it as the global context visual feature, because the entire image is passed through the encoder and the feature map at the last layer contain the holistic information about the image.
Although we use mask proposals to obtain the features only on masked regions on the feature map, these features already have comprehensive information about the scene.



\paragraph{Local-context Visual Features.}
\label{subsec:cropping}
To obtain local-context visual features given a mask proposal, we first mask the image and then crop the image to obtain a new image surrounding only an area of the mask proposal.
After cropping and masking the image, it is passed to the visual encoder of CLIP to extract our local-context visual feature $\mathbf{f}_m^L$:
\begin{align}
    \mathbf{f}_m^L=\phi_\text{CLIP}(\mathcal{T}_\text{crop}(I \odot m)),
    \label{eq:localvisfeat}
\end{align}
where $\mathcal{T}_{crop}(\cdot)$ denotes a cropping operation. 
This approach is commonly used in zero-shot semantic segmentation methods~\cite{ZSSS_1, ZSSS_2}.
Since this feature focuses on the masked region in the image where irrelevant regions are removed, it concentrates only on the target object itself.


\paragraph{Global-local Context Visual features.}

We aggregate global- and local-context features over masked regions to obtain one single visual feature that describe a representation of masked regions of the image.
The global-local context visual feature is computed as follows:
\begin{align}
    \mathbf{f}_m = \alpha ~\mathbf{f}^G_m + (1-\alpha)~  \mathbf{f}^L_m,
\end{align}
where $\alpha\in[0,1]$ is a constant parameter, $m$ is a mask proposal, $\mathbf{f}^G$ and $\mathbf{f}^L$ are global-context and local-context visual features in Eq.~\eqref{eq:globalvisfeat} and Eq.~\eqref{eq:localvisfeat}, respectively.
As in Eq.~\eqref{eq:sim}, the score for each mask proposal is then obtained by computing similarity between  our global-local context visual features and the textual feature of the expression described in the next section.


\subsection{Global-local Textual Features}
\label{subsec:textfeat}

Similar to the visual features, it is important to understand a holistic meaning as well as the target object noun in given expressions.
Given a referring expression $T$, we extract a global sentence feature, $\mathbf{t}^G$, using the pre-trained CLIP text encoder, $\psi_\text{CLIP}$, as follows:
\begin{align}
    \mathbf{t}^G=\psi_{\text{CLIP}}(T).
    \label{eq:globaltextfeat}
\end{align}
Although the CLIP text encoder can extract the textual representation aligning with the image-level representation, it is hard to focus on the target noun in the expression because the expression of this task is formed as a complex sentence containing multiple clauses, \eg \textit{``a dark brown leather sofa behind a foot stool that has a laptop computer on it"}.

To address this problem, we exploit a dependency parsing using spaCy~\cite{spacy} to find the target noun phrase, NP($T$), given the text expression $T$. 
To find the target noun phrase, we first find all noun phrases in the expression, and then select the target noun phrase that contains the root noun of the sentence. 
After identifying the target noun phrase in the input sentence, we extract the local-context textual feature from the CLIP textual encoder:
\begin{align}
    \mathbf{t}^L=\psi_\text{CLIP}(\text{NP}(T)).
    \label{eq:localtextfeat}
\end{align}

Finally, our global-local context textual feature is computed by a weighted sum of the global and local textual features described in Eq.~\eqref{eq:globaltextfeat} and Eq.~\eqref{eq:localtextfeat} as follows:
\begin{align}
    \mathbf{t} = \beta ~\mathbf{t}^G + (1-\beta)~  \mathbf{t}^L,
\end{align}
where $\beta\in[0,1]$ is a constant parameter, $\mathbf{t}^G$ and $\mathbf{t}^L$ are global sentence and local noun-phrase textual features, respectively.

\section{Implementation Details}
\label{sec:implementation}


We use unsupervised instance segmentation methods, FreeSOLO~\cite{Freesolo}, to obtain mask proposals, and the shorter size of an input image is set to 800.
For CLIP, the size of an image is set to 224x224.
The number of masking layers, $k$ in ViT is set to 3.
We set $\alpha=0.85$ for RefCOCOg, 0.95 for RefCOCO and RefCOCO+, and $\beta=0.5$ for all datasets.

\begin{table*}[ht!]
\footnotesize
\centering
    \caption{Comparison with Zero-shot RIS baseline methods on three standard benchmark datasets. U: The UMD partition. G: The Google partition. All baseline methods use FreeSOLO as the mask proposal network. $\dag$ denotes that the model is initialized with the ImageNet pre-trained weights and trained on RIS datasets. FreeSOLO upper-bound is computed between the GT mask and the maximum overlapped FreeSOLO mask with the GT mask.}

\begin{tabular}{c|ll|ccc|ccc|ccc}
\hline
\multirow{2}{*}{Metric} & \multirow{2}{*}{Methods}            & \multirow{2}{*}{\begin{tabular}[c]{@{}l@{}}Visual\\ Encoder\end{tabular}} & \multicolumn{3}{c|}{RefCOCO}                     & \multicolumn{3}{c|}{RefCOCO+}                    & \multicolumn{3}{c}{RefCOCOg}                     \\ \cline{4-12} 
& &     & val            & test A         & test B         & val            & test A         & test B         & val(U)            & test(U)        & val(G)         \\ \hline
\multirow{10}{*}{oIoU} & Supervised SoTA method~\cite{lavt} &                                                                           &    72.73            &    75.82            &   68.79             &  62.14              &   68.38             &    55.10            &     61.24           &    62.09            &   60.50             \\ \cline{2-12}
& \textbf{Zero-Shot Baselines} &                                                                           &                &                &                &                &                &                &                &                &                \\
& Grad-CAM                     & ResNet-50                                                                 & 14.02          & 15.07          & 13.49          & 14.46          & 14.97          & 14.04          & 12.51          & 12.81          & 12.86          \\
& Score map                     & ResNet-50                                                                 & 19.87          & 19.31          & 20.22          & 20.37          & 19.65          & 20.75          & 18.88          & 19.16          & 19.15          \\
& Region token                & ViT-B/32                                                                  & 21.71          & 20.31          & 22.63          & 22.61          & 20.91          & 23.46          & 25.52          & 25.38          & 25.29          \\
& Cropping                    & ResNet-50                                                                 & 22.36          & 20.49          & 22.69          & 23.95          & 22.03          & 23.49          & 28.20          & 27.64          & 27.47          \\
& Cropping                      & ViT-B/32                                                                  & 22.73          & 21.11          & 23.08          & 24.09          & 22.42          & 23.93          & 28.69          & 27.51          & 27.70          \\ \cline{2-12}
& Global-Local CLIP (ours)                         & ResNet-50                                                                 & 24.58          & 23.38          & 24.35          & 25.87          & 24.61          & 25.61          & 30.07          & 29.83          & 29.45          \\
& Global-Local CLIP (ours)                         & ViT-B/32                                                                  & \textbf{24.88} & \textbf{23.61} & \textbf{24.66} & \textbf{26.16} & \textbf{24.90} & \textbf{25.83} & \textbf{31.11} & \textbf{30.96} & \textbf{30.69} \\ 
\cdashline{2-12}[1pt/1pt]
& FreeSOLO upper-bound         & \multicolumn{1}{c|}{-}                                                    & 42.08          & 42.52          & 43.52          & 42.17          & 42.52          & 43.80          & 48.81          & 48.96          & 48.49          \\ 
\hline
\hline
\multirow{11}{*}{mIoU} & \textbf{Zero-Shot Baselines} &                                                                           &                &                &                &                &                &                &                &                &                \\
& Grad-CAM                    & ResNet-50                                                                 & 14.22          & 15.93          & 13.18          & 14.80         & 15.87          & 13.78        & 12.47          & 13.16          & 13.30          \\
& Score map                    & ResNet-50                                                                 & 21.32          & 20.96          & 21.57          & 21.61          & 21.17          & 22.30          & 20.07          & 20.43          & 20.63          \\
& Region token               & ViT-B/32                                                                  & 23.43          & 22.07          & 24.62          & 24.51          & 22.64          & 25.37          & 27.57          & 27.34          & 27.69          \\
& Cropping                 & ResNet-50                                                                 & 24.31          & 22.37          & 24.66          & 26.31          & 23.94          & 25.69          & 31.27          & 30.87          & 30.78          \\
& Cropping                 & ViT-B/32                                                                  & 24.83          & 22.58          & 25.72          & 26.33          & 24.06          & 26.46          & 31.88          & 30.94          & 31.06          \\ \cline{2-12}
& Global-Local CLIP (ours)            & ResNet-50                                                                 & \textbf{26.70} & \textbf{24.99} & 26.48          & \textbf{28.22} & \textbf{26.54} & \textbf{27.86} & 33.02          & 33.12          & 32.79          \\
& Global-Local CLIP (ours)            & ViT-B/32                                                                  & 26.20          & 24.94          & \textbf{26.56} & 27.80          & 25.64          & 27.84          & \textbf{33.52} & \textbf{33.67} & \textbf{33.61} \\ 
\cdashline{2-12}[1pt/1pt]
& FreeSOLO upper-bound                & \multicolumn{1}{c|}{-}                                                    & 48.25          & 46.62          & 50.43          & 48.28          & 46.62          & 50.62          & 52.44          & 52.91          & 52.76          \\ \cline{2-12}
& \textbf{Weakly-supervised method}   &                                                                           &                &                &                &                &                &                &                &                &                \\
& TSEG~\cite{tseg}                        & ViT-S/16$^\dagger$                                                        & 25.95          & -              & -              & 22.62          & -              & -              & 23.41          & -              & -              \\ \hline
\end{tabular}
\label{tab:total}

\end{table*}

\subsection{Masking in Global-context Visual Encoder}
\label{subsec:masking}
We use both ResNet-50 and ViT-B/32 architectures for the CLIP visual encoder.
Masking strategies of the global-context visual encoder for these two architecture are mostly similar but have small differences, described next. 

\paragraph{Masked Attention Pooling in ResNet~\cite{resnet}.}
In a ResNet-based visual encoder of the original CLIP, a global average pooling layer is replaced by an attention pooling layer.
This attention pooling layer has the same architecture as the multi-head attention in a Transformer.
A \textit{query} of the attention pooling layer is computed by a global average pooling operation onto the feature maps extracted by the ResNet backbone.
A \textit{key} and a \textit{value} of the attention pooling layer is given by a flattened feature map.
In our masked attention pooling, we mask the feature map using a given mask before computing \textit{query}, \textit{key} and \textit{value}.
After masking feature maps, we compute \textit{query}, \textit{key} and \textit{value}, and then they are fed into the multi-head attention layer.
The detailed illustration of our masked attention pooling in ResNet is shown in Figure~\ref{fig:global}a.


 
 
 

 
\paragraph{Token Masking in ViT~\cite{vit}.}
\label{para:token_masking}
Following ViT, we divide an image into grid patches, and embed patches to a linear layer with positional embeddings to get tokens, and then process those tokens with a series of Transformer layer.
To capture global-context of images, we mask tokens in only the last $k$ Transformer layers.
The tokens are reshaped and masked by a given mask proposal, and then flattened and applied to the subsequent Transformer layer.
As ViT has a class token (CLS), we use the final output feature from this CLS token as our global-context visual representation.
The detailed method of our token masking in ViT is also shown in Figure~\ref{fig:global}b.
In our experiments, we use ViT-B/32 architecture for the backbone of our ViT-based visual encoder, and we apply a token masking to the last 3 layers in the visual encoder.
We show the performances with respect to the location of token masking layers in the supplementary materials.


%

\section{Experiments}

%

\begin{figure}[t]
    \vspace{-0.4cm}
      \begin{subtable}[c]{0.5\linewidth}
        \scalebox{0.65}{
        \begin{tabular}{c|c|c|c}
        \hline
        &&\multicolumn{2}{c}{oIoU on PhraseCut} \\\cline{3-4} 
        Method & Train dataset & All & Unseen \\
                                 \hline
        \multirow{3}{*}{CRIS} & RefCOCO & 15.53 & 13.75 \\
                                & RefCOCO+ & 16.30 & 14.62 \\
                                & RefCOCOg & 16.24 & 13.88 \\ \hdashline[1pt/1pt]
        \multirow{3}{*}{LAVT}    & RefCOCO                    & 16.68                   & 14.43                  \\
                                 & RefCOCO+                   & 16.64                   & 13.49                  \\
                                 & RefCOCOg                   & 16.05                   & 13.48                  \\ \hdashline[1pt/1pt]
        Ours                & N/A                  & \textbf{23.64}                   & \textbf{22.98}                  \\ \hline
        \end{tabular}
        }
      \label{tab:rebuttal:zero-shot}
      \end{subtable}
      \hspace{0.4cm}
      \begin{subfigure}[c]{0.4\linewidth}
      \vspace{0.4cm}
      \includegraphics[width=1\linewidth, right]{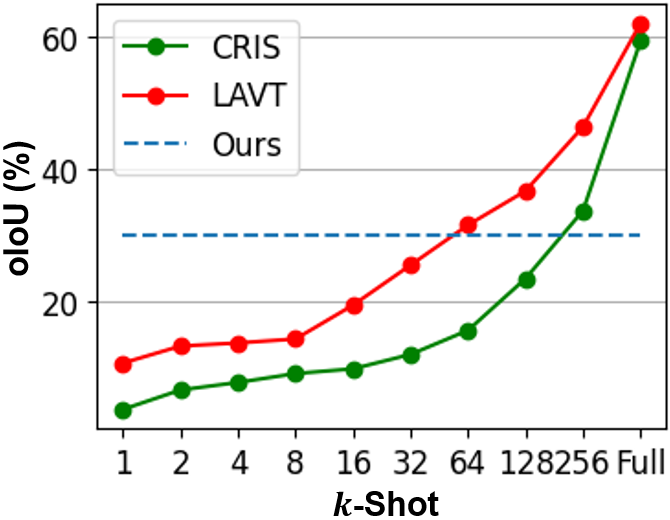} 
      \label{fig:few-shot}
      \end{subfigure}

      \vspace{-0.6cm}
    \caption{Comparisons to supervised methods in zero-shot setting on PhraseCut (left), and in few-shot setting on RefCOCOg (right). Unseen denotes a subset with classes that are not seen in RefCOCO.
    }
    \label{fig:comparison}
    \vspace{-0.3cm}
\end{figure}

\subsection{Datasets and Metrics}
We evaluate our method on RefCOCO~\cite{refcoco}, RefCOCO+~\cite{refcoco} and RefCOCOg~\cite{refcocog, refcocog_google}, where the images and masks in MS-COCO~\cite{coco} dataset are used to annotate the ground-truth of the referring image segmentation task.
RefCOCO, RefCOCO+ and RefCOCOg have 19,994, 19,992 and 26,711 images with 142,210, 141,564 and 104,560 referring expressions, respectively.
RefCOCO and RefCOCO+ have shorter expressions and an average of 1.6 nouns and 3.6 words are included in one expression, while RefCOCOg expresses more complex relations with longer sentences and has an average of about 2.8 nouns and 8.4 words.
The detailed statistics of those datasets are demonstrated in our supplementary materials.


\begin{table}[]
\footnotesize
\centering
\caption{oIoU results of our method and the baselines using COCO instance GT masks. We use a ViT-B/32 model for a visual encoder.}
\begin{tabular}{l|ccc}
\hline
Method  & RefCOCO & RefCOCO+ & RefCOCOg \\ \hline
Grad-CAM    & 18.32 & 18.14 & 21.24  \\
Score map & 23.97 & 25.50 & 28.11   \\
Region token & 35.59 & 38.13 & 40.19 \\
Cropping & 36.32 & 42.07 & 47.42   \\ \hdashline[1pt/1pt]
Ours   & \textbf{37.05} & \textbf{42.59} & \textbf{51.01} \\ \hline
\end{tabular}
\label{tab:cocogt}
\end{table}


\begin{table}[]
\centering
\footnotesize
\caption{oIoU results with different context-level features on the val split of RefCOCOg. We use a ViT-B/32 model for a visual encoder.}
\begin{tabular}{l|l|ccc}
\hline
\multicolumn{2}{c|}{\multirowcell{2.2}{Encoder Variants}}  &   \multicolumn{3}{c}{Textual features} \\
\cline{3-5}
 \multicolumn{2}{c|}{} & Global         & Local         & Global-Local  \\ \hline
\multirow{3}{*}{\makecell{ Visual \\ features }}& Global        &  27.03             &  {27.37}   &     \underline{27.60}\\
& Local         &  {28.69}    &  25.23            &     \underline{29.48}         \\
& Global-Local        &  \underline{30.18}             &  \underline{27.94}            &     \textbf{31.11} \\
\hline
\end{tabular}
\label{tab:globallocal}
\end{table}

For the evaluation metrics, we use the overall Intersection over Union (oIoU) and the mean Intersection over Union (mIoU) which are the common metrics for the referring image segmentation task.
The oIoU is measured by the total area of intersection divided by the total area of union, where the total area is computed by accumulating over all examples.
In our ablation study, we use oIoUs since most of supervised RIS methods~\cite{restr, vlt} adopt it.
We also report the mIoUs as in \cite{tseg}, which computes the average IoU across all examples while considering the object sizes.

    \begin{figure}
        \centering
        \includegraphics[width=0.85\linewidth]{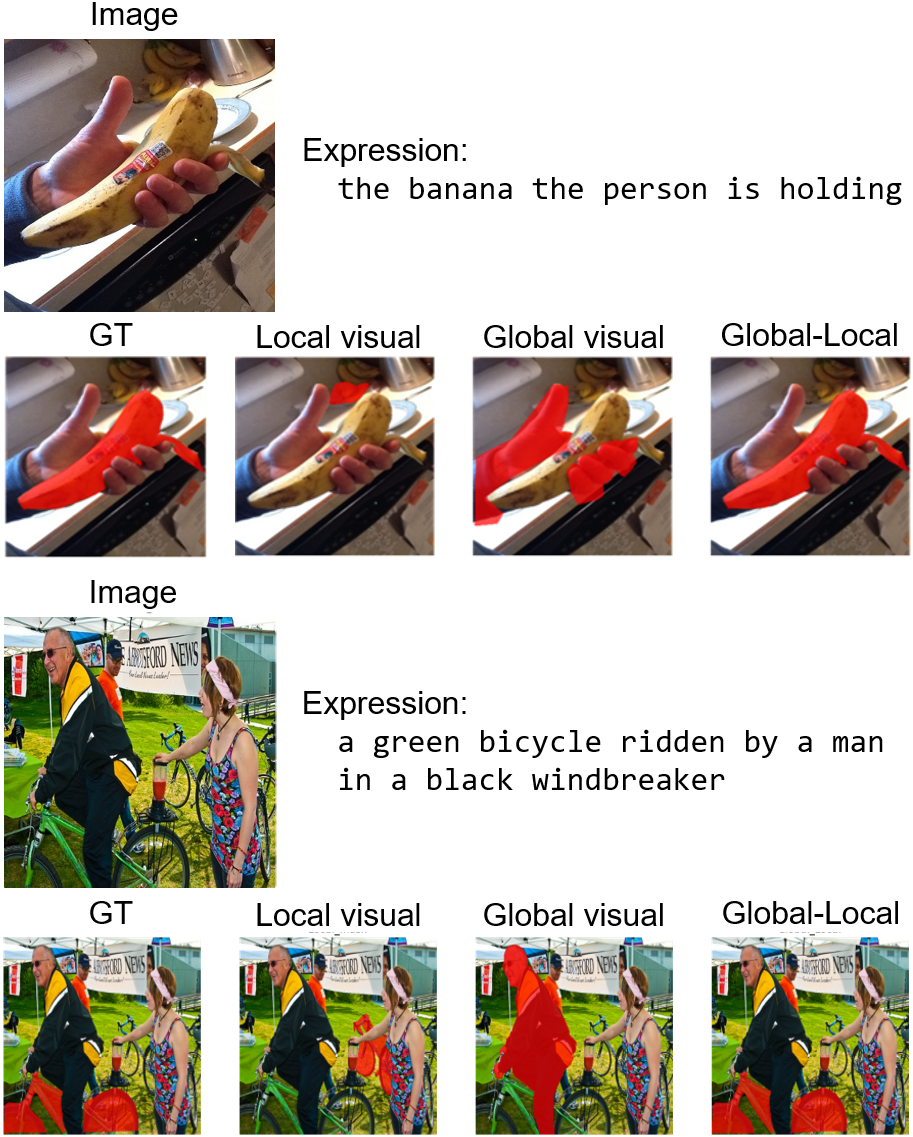}
        \caption{Qualitative results with different levels of visual features. COCO instance GT masks are used as mask proposals to validate the effect of the global-local context visual features.}
        \label{fig:qual_visual}
    \end{figure}

\subsection{Baselines}


We modify some baseline methods extracting dense predictions from CLIP into zero-shot RIS task to compare with our framework, and use FreeSOLO~\cite{Freesolo} as a mask generator in all baselines.

\begin{itemize}[leftmargin=*]
    \item \textbf{Grad-CAM:} The first baseline is utilizing gradient-based activation map based on Grad-CAM~\cite{gradcam} which has been verified in the prior work~\cite{clipcam}.
    After obtaining the activation maps using the similarity score of image-text pairs, we mask the maps and aggregate scores for all mask proposals, and select the mask with the highest score.
    %
    
    \item \textbf{Score Map:} The second baseline is the method extracting a dense score map as in MaskCLIP~\cite{MaskCLIP}. 
    As in MaskCLIP, to obtain dense score maps without pooling, a \textit{value} linear layer and the last layer in the attention pooling are transformed into two consecutive 1$\times$1 convolution layers. 
    The feature map extracted from ResNet is forwarded to those two layers to get language-compatible dense image feature map, and then compute a cosine similarity with CLIP's textual feature.
    After obtaining a score map, we project mask proposals to a score map. 
    The scores in the mask area are averaged and then we select the mask with the maximum score.
    
    \item \textbf{Region Token in ViT:} The third baseline is a method used in Adapting CLIP~\cite{adaptingclip}. 
    Similar to Adapting CLIP, we use region tokens for each mask proposal for all Transformer layers in CLIP's visual encoder instead of using superpixels.
    We finally compute the cosine similarity between each class token of a mask proposal and CLIP's textual feature, and then choose the mask with the highest score.

    
    

    \item \textbf{Cropping:} The last baseline is our local-context visual features described in Section~\ref{subsec:cropping}. 
    Cropping and masking is a commonly used approach utilizing CLIP for extracting mask or box region feature in a range of zero-shot dense prediction tasks~\cite{OVOD_ViLD, OVOD_prompt, ZSSS_1, ZSSS_2, reclip}. 
    Therefore, we consider cropping as one of the zero-shot RIS baselines.
    
\end{itemize}



    \begin{figure}
        \centering
        \includegraphics[width=1\linewidth]{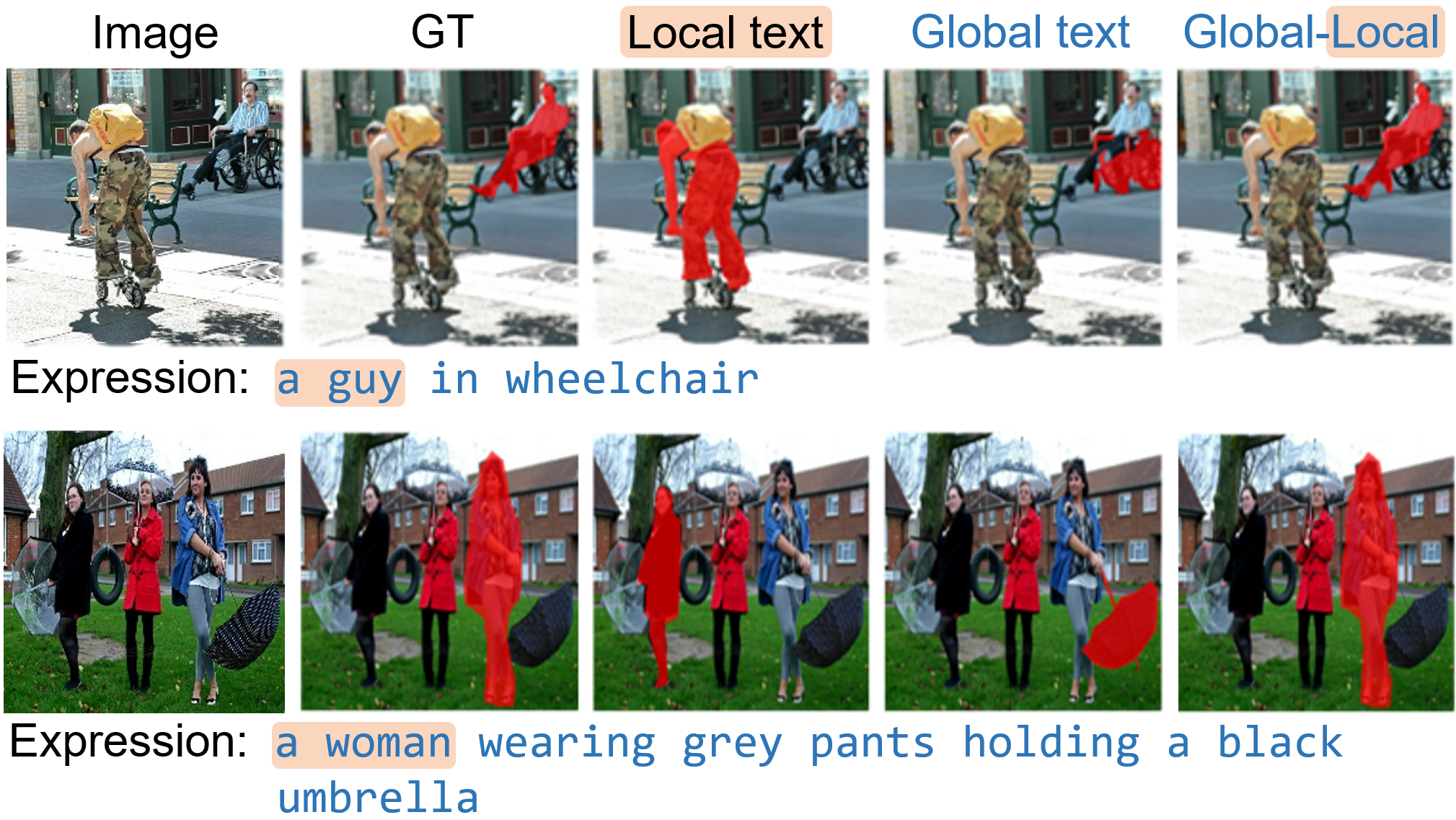}
        \caption{Qualitative results with different levels of textual features using COCO Instance GT mask proposals. 
        }
        \label{fig:qual_text}
    \end{figure}

\subsection{Results}
\paragraph{Main Results.}
We report referring image segmentation performances of our global-local CLIP and other baselines on RefCOCO, RefCOCO+ and RefCOCOg in terms of oIoU and mIoU metrics in Table~\ref{tab:total}.
For a fair comparison, all methods including baselines use FreeSOLO~\cite{Freesolo} mask proposals to produce the final output mask.
The experimental results show that our method outperforms other baseline methods with substantial margins.
Our method also surpasses the weakly supervised referring image segmentation method (TSEG)~\cite{tseg} in terms of mIoU\footnote{We only compare mIoU scores with TSEG since it reports only mIoU scores in the paper.}.
We also show upper-bound performances of using FreeSOLO, where the scores are computed by the IoU between ground-truth masks and its max-overlapped mask proposal. 
Although there is still a gap compared to the fully-supervised referring image segmentation methods, our method improves performances significantly compared to the baselines with the same upper-bound.

\paragraph{Zero-shot Evaluation on Unseen Domain.}
To verify the effectiveness of our method in a more practical setting, we report the zero-shot evaluation results with SoTA supervised methods~\cite{lavt, cris} on the test split of PhraseCut~\cite{phrasecut} in Figure~\ref{fig:comparison} (left).
Note that, RefCOCO contains expressions for only 80 salient object classes, whereas PhraseCut covers a variety of additional visual concepts \ie 1272 categories in the test set.
Our method outperforms both supervised methods, even though our models were never trained under RIS supervision. 
When evaluated on a subset of classes that are not seen in the RefCOCO datasets (\textit{Unseen} column), the supervised methods show significant performance degradation, whereas our method works robustly on this subset.

\paragraph{Comparison to supervised methods in few-shot Setting.}
We also compare our model to two supervised RIS methods~\cite{lavt, cris} in a few-shot learning setting, where the training set includes $k$ instances for each of 80 {classes in RefCOCO}\footnote{{we use object classes in RefCOCO GT annotation. This is to cover all salient objects in the dataset during the few-shot training.}}.
Note that the supervised methods use additional forms of supervision in training, whereas our method does not require any form of training or additional supervision; thus this setting is even disadvantageous to our method.
Figure~\ref{fig:comparison} (right) shows oIoU while varying $k$ on RefCOCOg.
The results clearly show that our method outperforms both supervised methods with large margins when $k$ is small, and the gaps narrow as $k$ gets larger (64 and 256 for LAVT~\cite{lavt} and CRIS~\cite{cris}, respectively).
Note that it covers about 10\% of the training set when $k=64$ and the same trends hold for both RefCOCO and RefCOCO+.


\subsection{Ablation Study}
\label{subsec:ablation}

    \begin{figure*}[t]
        \centering
        \includegraphics[width=0.95\linewidth]{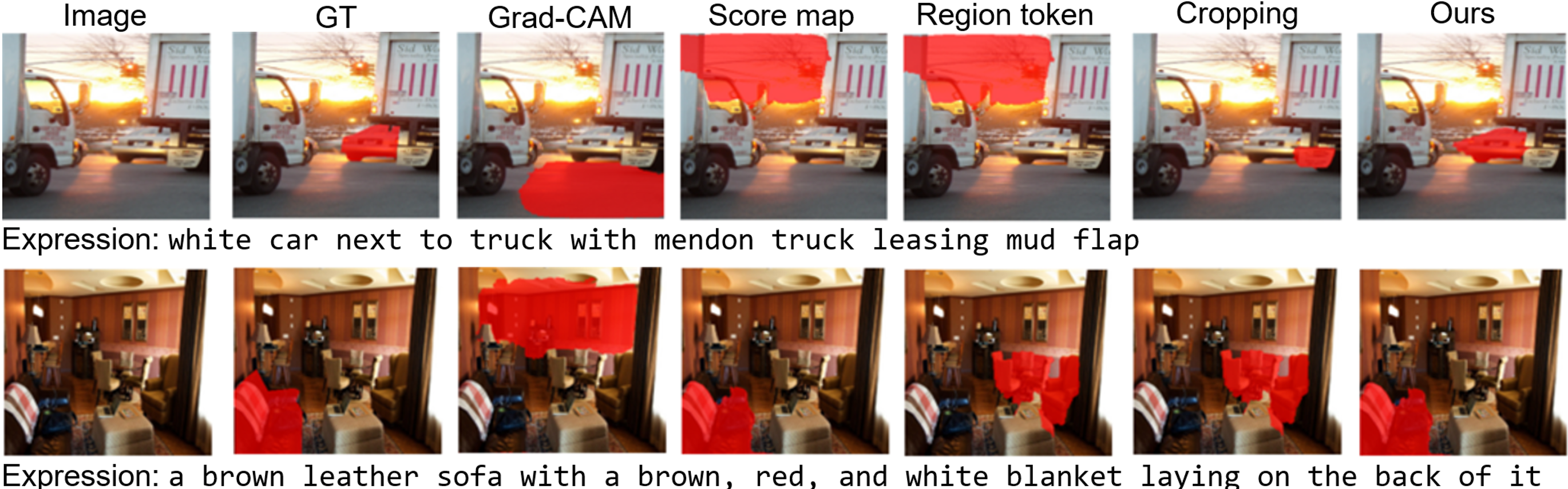}
        \caption{Qualitative results of our method with the several baselines. Note that all methods use mask proposals generated by FreeSOLO.}
        \label{fig:qualitative}
    \end{figure*}

\paragraph{Effects of Mask Quality.}
To show the impact of the proposed method without considering the mask quality of the mask generators, we evaluate the performance of our method and the baselines with COCO instance GT masks in Table~\ref{tab:cocogt}.
Our approach has demonstrated superior performance compared to all baselines and has shown a performance improvement of over 3.5\%, particularly on RefCOCOg which includes longer expressions. 
We believe that our method performs well on challenging examples that involve complex expressions, such as those with multiple clauses, which require an understanding of both the language and the scene.


\paragraph{Effects of Global-Local Context Features.}

We also study the effects of global-local context features in both visual and textual modalities and show the results in Table~\ref{tab:globallocal}.
For this analysis, we use RefCOCOg as it contains more complex expressions with multiple clauses.
Among all combinations of two modalities, using both global-local context features in the visual and textual domains leads to the best performance.

\paragraph{Qualitative Analysis.}


We demonstrate several results that support the effectiveness of our global-local context visual features in Figure~\ref{fig:qual_visual}.
To show this effect more clearly, we use COCO instance GT masks as mask proposals. 
When using only local-context visual features, the predicted mask tends to focus on the instance that shares the same class as the target object. 
However, when using only global-context visual features, the predicted mask tends to capture the context of the expression but may focus on a different object class.
By combining global and local context, our method successfully finds the target mask.
We also demonstrate the effectiveness of our global-local context textual features in Figure~\ref{fig:qual_text}.
Furthermore, we compare the qualitative results of our method with baseline methods in Figure~\ref{fig:qualitative}. Our proposed global-local CLIP outperforms the baseline methods in identifying the target object by taking into account the global context of the image and expression.

\label{sec:experiments}
\section{Conclusion}
\label{sec:conclusion}

In this paper, we propose a simple yet effective zero-shot referring image segmentation framework focusing on transferring knowledges from image-text cross-modal representations of CLIP.
To tackle the difficulty of the referring image segmentation task, we propose global-local context encodings to compute similarities between images and expressions, where both target object semantics and relations between the objects are dealt in a unified framework.
The proposed method significantly outperforms all baseline methods and weakly supervised method as well.
\vspace{-0.2cm}
\small{
\paragraph{Acknowledgement.}
This work was supported by the IITP grants (No.2019-0-01842, No.2021-0-02068, No.2022-0-00926) funded by MSIT, the ISTD program (No.20018334) funded by MOTIE, and the GIST-MIT Research Collaboration grant funded by GIST, Korea.
}

{\small
\bibliographystyle{ieee_fullname}
\bibliography{11_references}
}

\ifarxiv
\clearpage
{\noindent\large\textbf{Supplemental materials}}
\appendix
\label{sec:appendix}
\setcounter{figure}{7}
\setcounter{table}{3}



\section{Analysis on Global-local Textual Feature}

\paragraph{Dataset statistics.}
The datasets used in our paper, RefCOCO, RefCOCO+ and RefCOCOg, have difference characteristics.
As we mentioned in the original manuscript, RefCOCO and RefCOCO+ have shorter expressions and an average of 1.6 nouns and 3.6 words are included in one expression, while RefCOCOg expresses more complex relations with longer sentences and has an average of about 2.8 nouns and 8.4 words. 
In this supplementary material, we further analyze frequencies with respect to the number or words and nouns in the sentence.
As shown in Figure~\ref{fig:statistic_dataset}, RefCOCOg contains much longer expressions than two other datasets.


    \begin{figure}[h]
        \centering
        \includegraphics[width=1\linewidth]{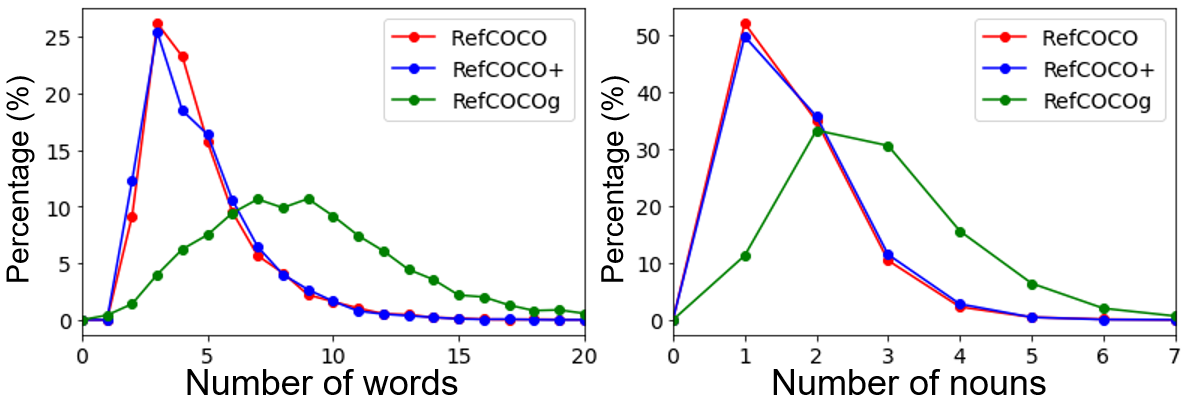}
        \caption{Statistics of datasets. We investigate the number of words and the number of nouns in the sentence on each datasets.}
        \label{fig:statistic_dataset}
    \end{figure}

\paragraph{Effects of global-local textual features.}
Our global-local context textual feature, which is designed to focus on the target noun phrase as well as the whole sentence, is highly effective in longer sentences.
To analyze effects of our method with respect to the length of sentences, we show oIoU differences between global-local context textual features and global-context textual features with respect to the length of sentences in Figure~\ref{fig:oiou_difference}.
It clearly shows that the performances using global-local textual features outperform global-context textual features when the expressions contain more words.

As we illustrated in Figure~4 in the original manuscript, there are no significant difference between global textual features and global-local textual features in RefCOCO and RefCOCO+ datasets. 
We found that RefCOCO and RefCOCO+ contains shorter sentences where the extracted noun phrase is the same as the sentence.
If the target noun phrase is the same as the sentence, our global-local context textual features have no advantage over the global-context one.
We measures the percentage of the case that the target noun phrase is the same as the sentence, and there are 48.61\% on RefCOCO, 43.43\% on RefCOCO+ and 7.46\% on RefCOCOg.
This explains that the lower performance gains of using our global-local textual features on RefCOCO and RefCOCO+ datasets.




    \begin{figure}[t]
        \centering
        \includegraphics[width=0.9\linewidth]{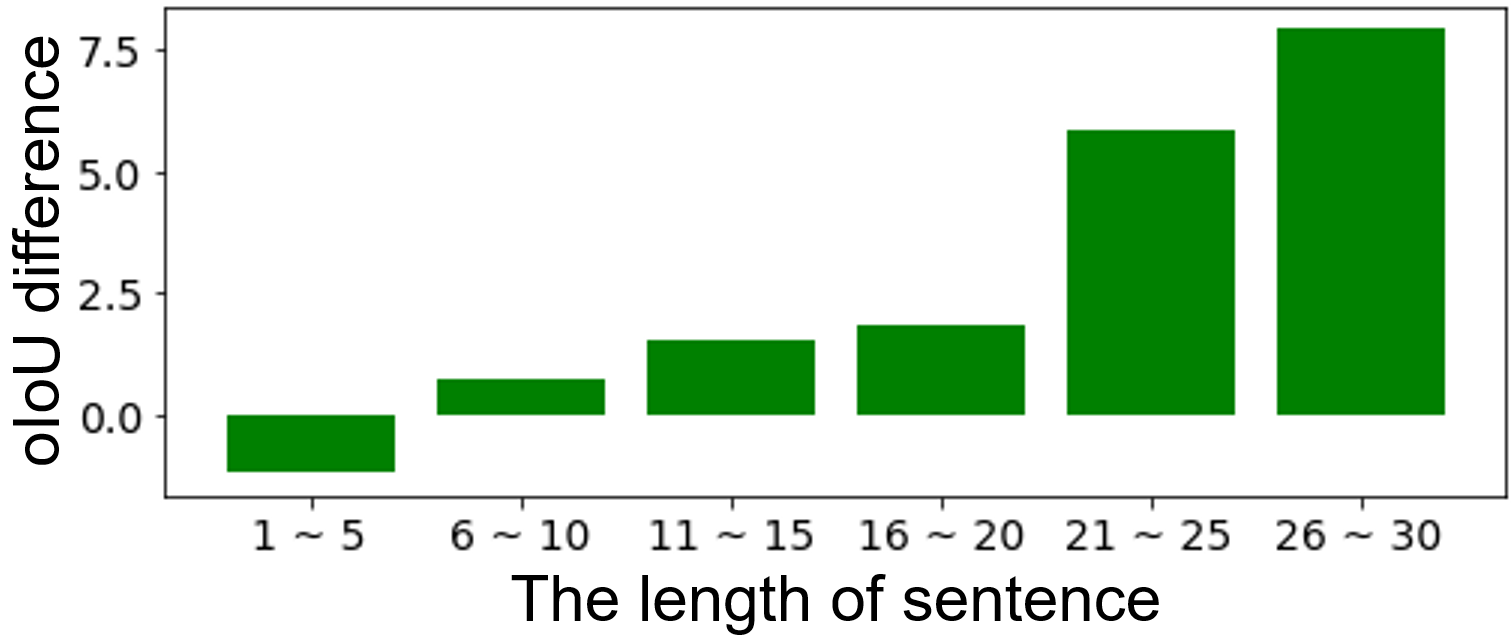}
        \caption{oIoU differences between global-local context textual feature and global context textual feature with respect to  length of sentences. We conduct the experiment on RefCOCOg dataset, since this has much longer expressions than two other datasets.}
        \label{fig:oiou_difference}
    \end{figure}


    
\begin{table}[t]
\centering{}
\small

\caption{Examples of the \colorbox{pink}{target noun phrases} in the sentences}
\begin{tabular}{l}
\toprule
Examples: \\
\midrule
\hspace{0.2cm} \colorbox{pink}{mom} \\
\hspace{0.2cm} \colorbox{pink}{little girl} \\
\hspace{0.2cm} near \colorbox{pink}{zebra} \\
\hspace{0.2cm} right \colorbox{pink}{sandwich} \\
\hspace{0.2cm} \colorbox{pink}{girl's umbrella} \\
\hspace{0.2cm} \colorbox{pink}{glass} of juice in table \\
\hspace{0.2cm} \colorbox{pink}{yellow baked squash dish} \\
\hspace{0.2cm} left \colorbox{pink}{person} with elbow bent \\
\hspace{0.2cm} \colorbox{pink}{child} sitting on womans lap \\
\hspace{0.2cm} \colorbox{pink}{a cow's ear} with a circular tag \\
\hspace{0.2cm} flowered \colorbox{pink}{quilt} on back of couch \\
\hspace{0.2cm} \colorbox{pink}{a mother giraffe} licking her baby \\
\hspace{0.2cm} with bruises! okey, \colorbox{pink}{closest ugly couch} \\
\hspace{0.2cm} \colorbox{pink}{a black and white dog} with pointy ears \\
\hspace{0.2cm} that was it ... \colorbox{pink}{man} in the center up front \\
\hspace{0.2cm} \colorbox{pink}{the baby boy} wearing a red shirt and gray bib \\
\hspace{0.2cm} \colorbox{pink}{a flat box} full of plants labeled wegman's nursery \\
\hspace{0.2cm} \colorbox{pink}{a man's black tie} under all the other ties he is wearing \\
\bottomrule
\label{tab:noun_phrase}
\end{tabular}
\end{table}

\paragraph{Target noun phrase extraction using SpaCy~\cite{spacy}.}
To extract local-context textual feature, we need find the target noun phrase in a whole sentence.
In this supplementary material, we describe the detailed process of obtaining the target noun phrase using a dependency parsing tool, SpaCy, as follows.
SpaCy extracts all noun phrases and the root word of a sentence.
The root word, which is also referred to as a head word in SpaCy, is the word that has no dependency with other words, \ie the word that does not have the parent word in the dependency tree.
If the root word is a verb, we use the root word's children noun as the root word, and then select the noun phrase containing the root word.
If there is no noun phrase containing the root word, we use the whole sentence as the target noun phrase.
We show examples of the target noun phrase extracted from the expression in table~\ref{tab:noun_phrase}.






    \begin{figure*}[t]
        \centering
        \begin{subfigure}[t]{0.32\textwidth}
            \centering
            \includegraphics[width=1\linewidth]{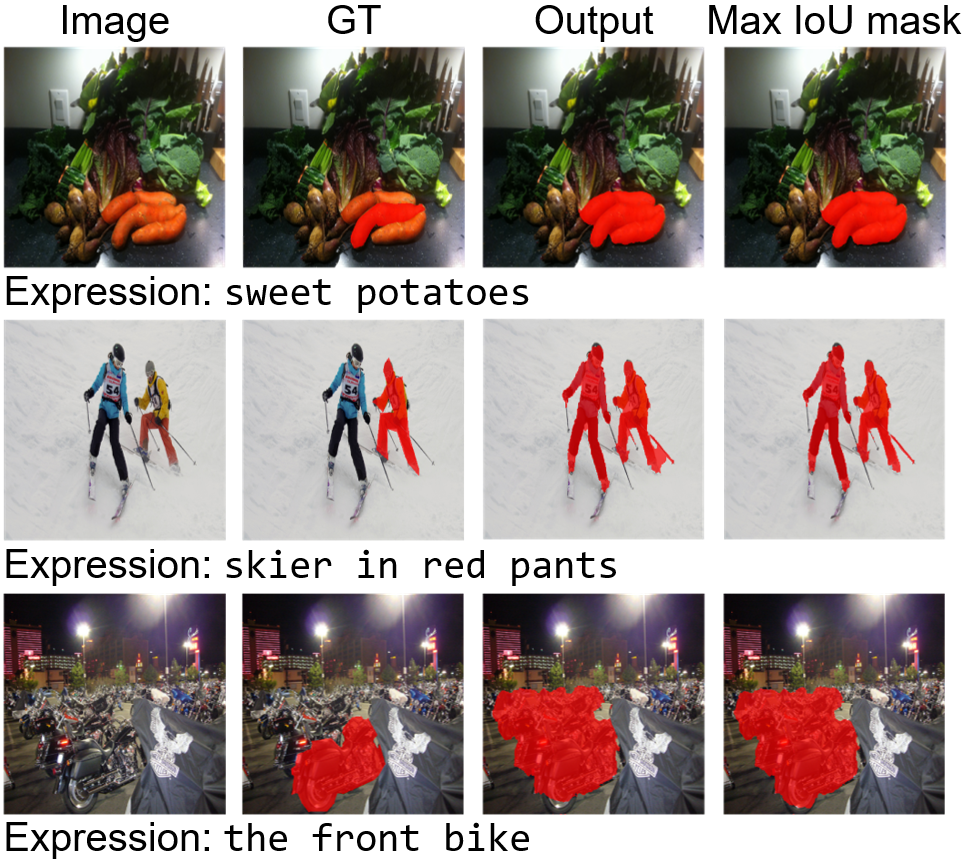}
            \caption{Quality of mask proposals}
            \label{subfig:failure_case_a}
        \end{subfigure}
        \hfill
        \begin{subfigure}[t]{0.32\textwidth}
            \centering
            \includegraphics[width=1\linewidth]{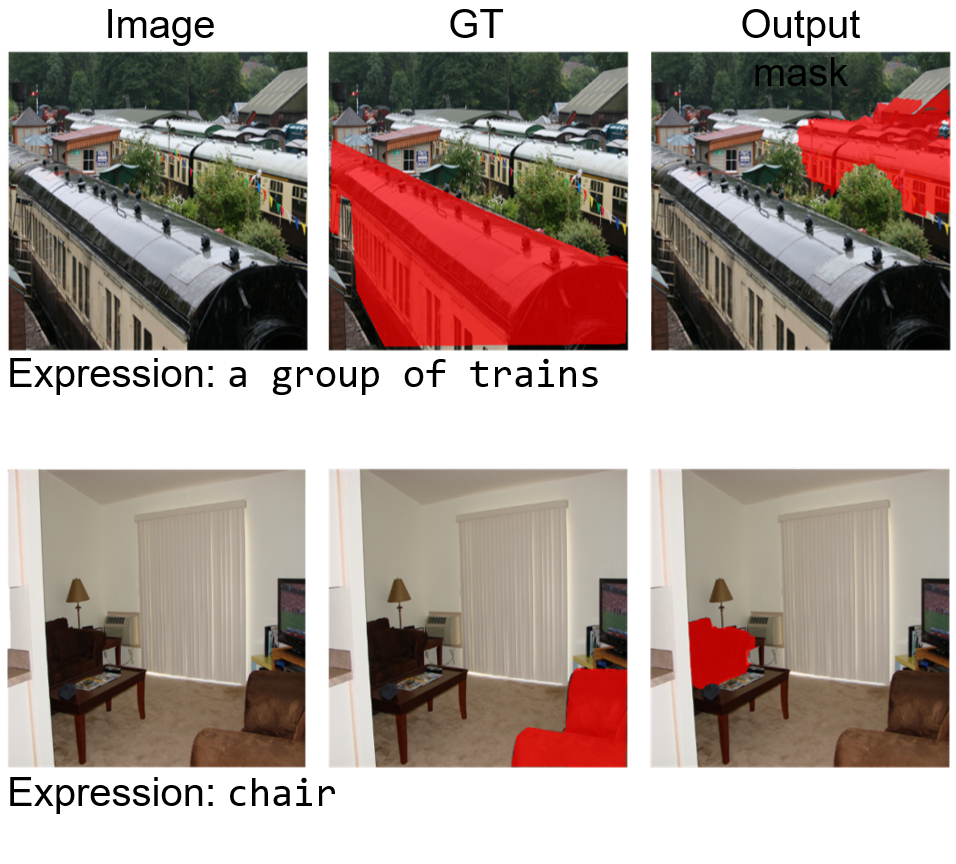}
            \caption{Unclear expressions and mask annotations}
            \label{subfig:failure_case_b}
        \end{subfigure}
        \hfill
        \begin{subfigure}[t]{0.32\textwidth}
            \centering
            \includegraphics[width=1\linewidth]{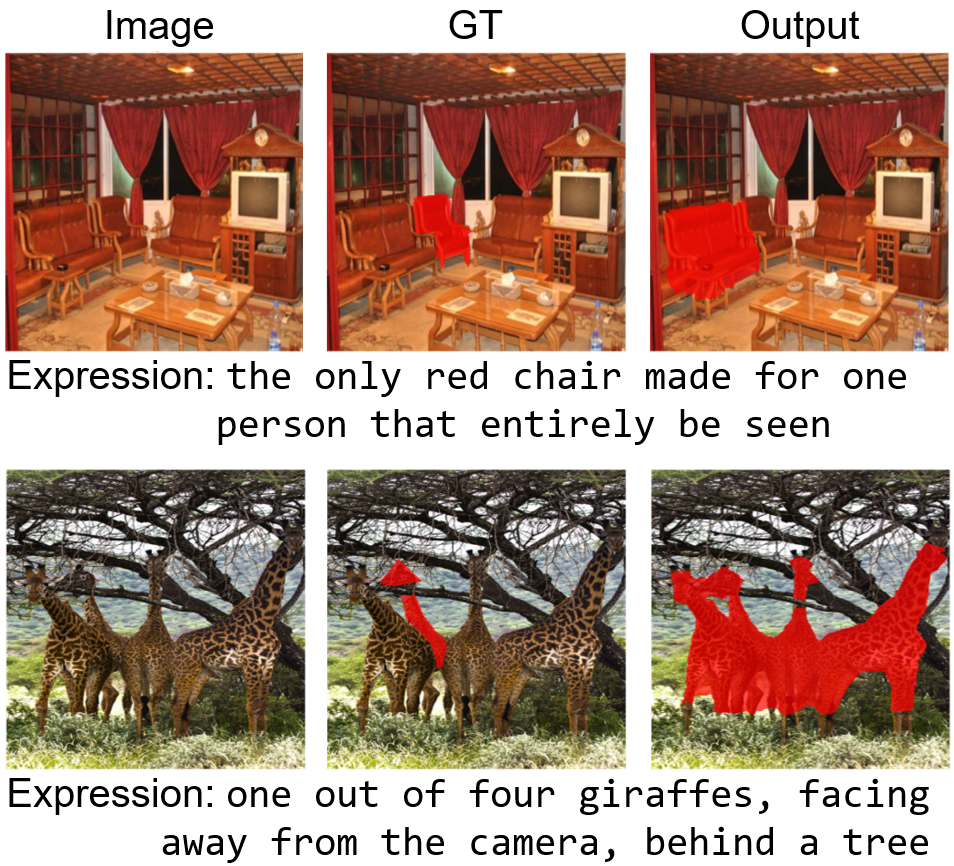}
            \caption{Complicated expressions}
            \label{subfig:failure_case_c}
        \end{subfigure}
        \hfill
        \caption{Failure cases. (a) Our method depends on the quality of mask proposals. Mask proposals generated by FreeSOLO are sometimes unable to distinguish multiple instances close together. (b) Some labeling noises are included in the datasets. There are multiple possible answers to the given expression. (c) Some expressions are too complex for CLIP to understand.}
        \label{fig:failure_cases_with_subfig}
        
    \end{figure*}

\section{Analysis on hyperparameters $\alpha$ and $\beta$.}
Our method uses two hyperparmaters, $\alpha$ and $\beta$, that combine global and local features for each modality.
We present oIoU scores with respect to these parameters for all datasets in Figure~\ref{fig:graph}.
For the visual features, we first fix $\beta=1$ and then choose $\alpha=0.85$ on RefCOCOg and $\alpha=0.95$ on RefCOCO and RefCOCO+.
With these parameters, we set $\beta=0.5$ for the text features.

    \begin{figure}[h]
        \centering
        \includegraphics[width=1\linewidth]{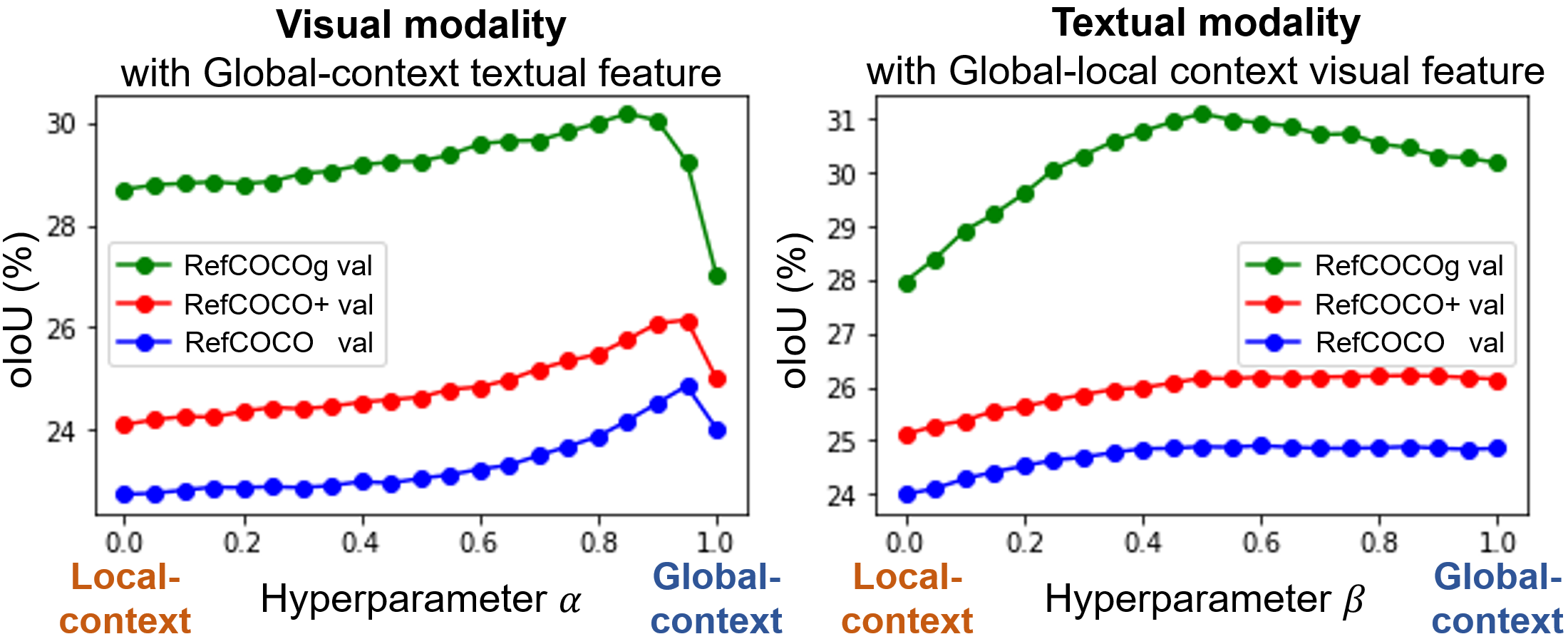}
        \caption{Analysis on hyperparameters $\alpha$ and $\beta$ for each visual and textual modality.}
        \label{fig:graph}
    \end{figure}

\begin{figure}[h]
    \centering
    \includegraphics[width=1\linewidth]{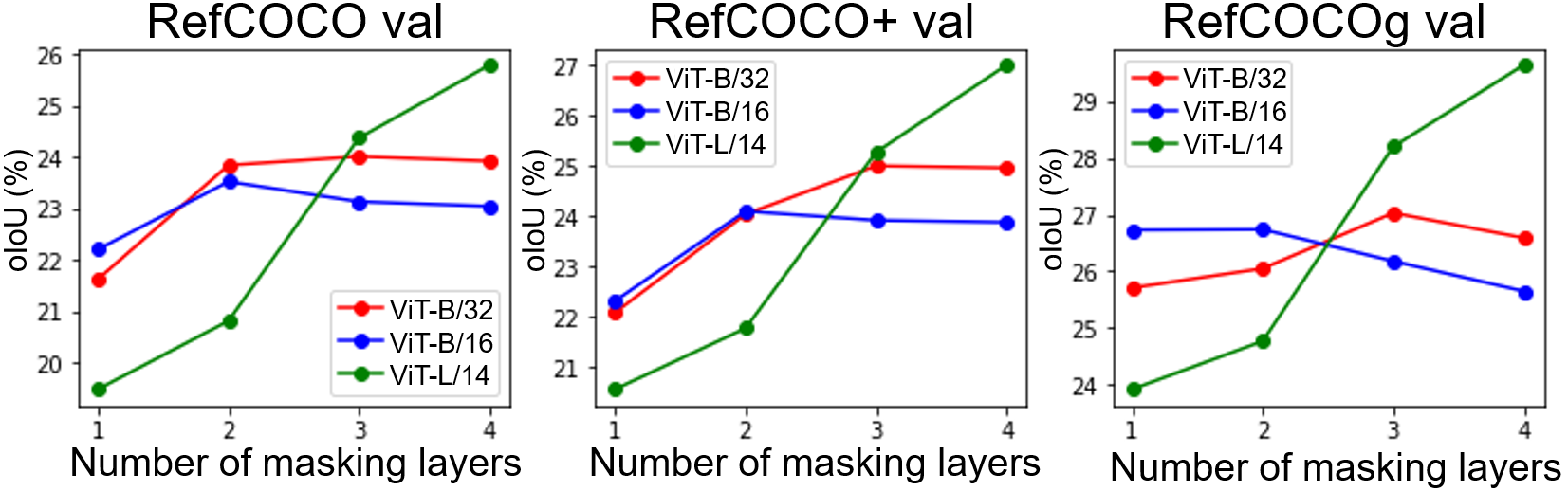}
    \caption{Ablation study on number of masking layers k from the last layer in ViT. }
    \label{fig:ablation_token_masking_graph}
\end{figure}

\section{Ablation Study on Token Masking in ViT}
Token masking is a method for extracting global-context visual features in ViT models.
We mask tokens in only the last k Transformer layers to capture global-context of images.
Figure~\ref{fig:ablation_token_masking_graph} shows the oIoU results with respect to the $k$-th token masking layers from the last layer and the ViT variants of CLIP.
We use global-context textual feature to compute the cosine similarity with visual feature and mask proposals from FreeSOLO~\cite{Freesolo}.
We choose $k$ $=$ 3 on ViT-B/32, $k$ $=$ 2 on ViT-B/16 and $k$ $=$ 4 on ViT-L/14, which show the best performances.




    \begin{figure}[t]
        \centering
        \includegraphics[width=0.9\linewidth]{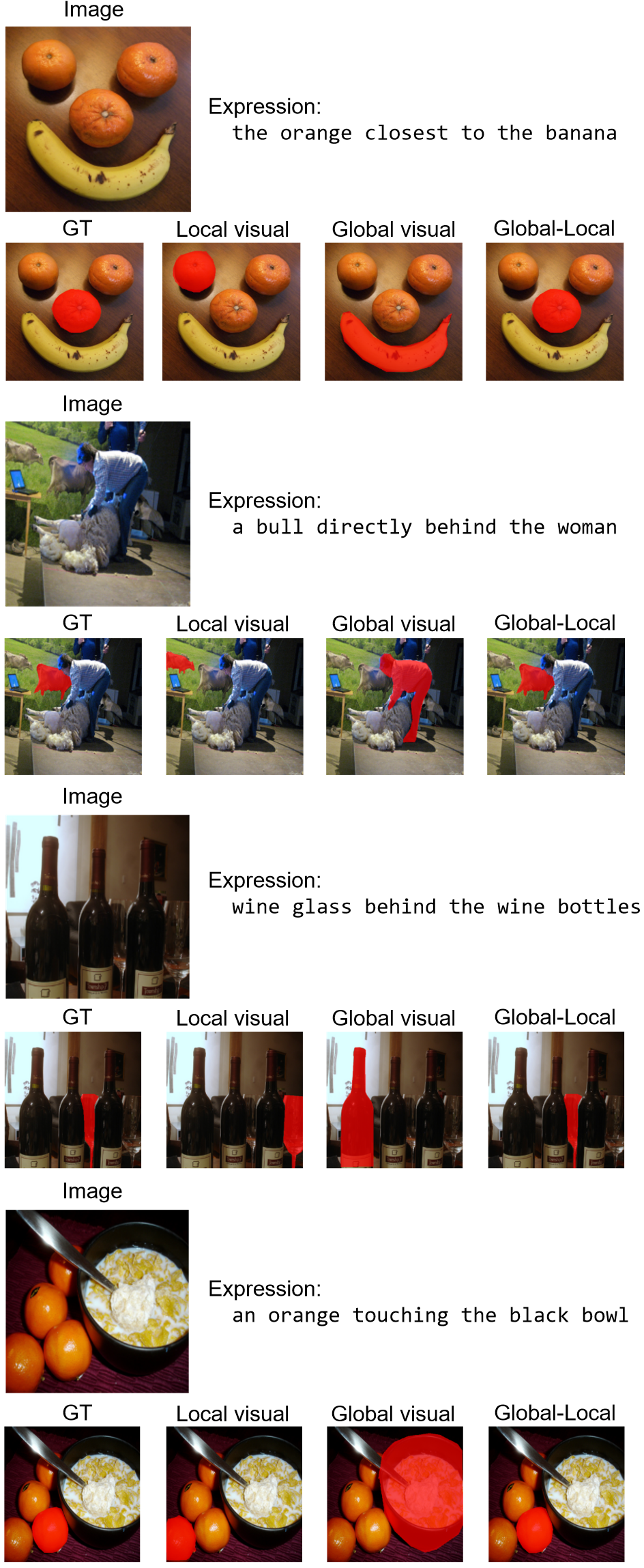}
        \caption{Additional qualitative results with different levels of visual features. COCO instance GT masks are used as mask proposals to validate an effect of the global-local context visual features.}
        \label{fig:suple_qual_visual}
    \end{figure}
    
\section{Additional Qualitative Results}

\paragraph{Failure cases.}

        

    
We show failure cases of our Global-Local CLIP in Figure~\ref{fig:failure_cases_with_subfig}.
We categorize failure cases into three groups: (a) failure cases due to the quality of FreeSOLO~\cite{Freesolo} (b) Unclear expression and mask annotations, and (c) too complicated expressions.
Since FreeSOLO, which we use to generate mask proposals, is a unsupervised instance segmentation framework, it does not utilize any category labels for training to generate class-agnostic instances masks. 
Thus it tends to find a cluster of the instances with the same semantic class because it cannot distinguish multiple instances close together.
We illustrate the maximum overlapped mask with ground-truth generated by FreeSOLO in the last column of Figure~\ref{subfig:failure_case_a}.
Moreover, there are a lot of labeling noises in the datasets.
Since the task of referring image segmentation is extremely difficult, it is hard to obtain clean annotations.
There are multiple possible answers in the image, but the ground-truth mask may contain only one instance as shown in Figure~\ref{subfig:failure_case_b}.
Furthermore, some expressions are too complex to understand with CLIP, because the text encoder of CLIP is not designed to handle complicated expressions.
Therefore our method also has limited ability of understand the complicated natural languages as shown in Figure~\ref{subfig:failure_case_c}.

\paragraph{More qualitative results.}

We demonstrate more results that support the effects of global-local visual and textual context features in Figure~\ref{fig:suple_qual_visual} and \ref{fig:suple_qual_text}.
As in Figure 5 and 6 in the original manuscript, we use COCO instance GT masks as mask proposals to show a clear impact of our global-local features.
We also illustrate more qualitative results of our method with the several baselines in Figure~\ref{fig:more_qualitative}.


    \begin{figure}[t]
        \centering
        \includegraphics[width=1\linewidth]{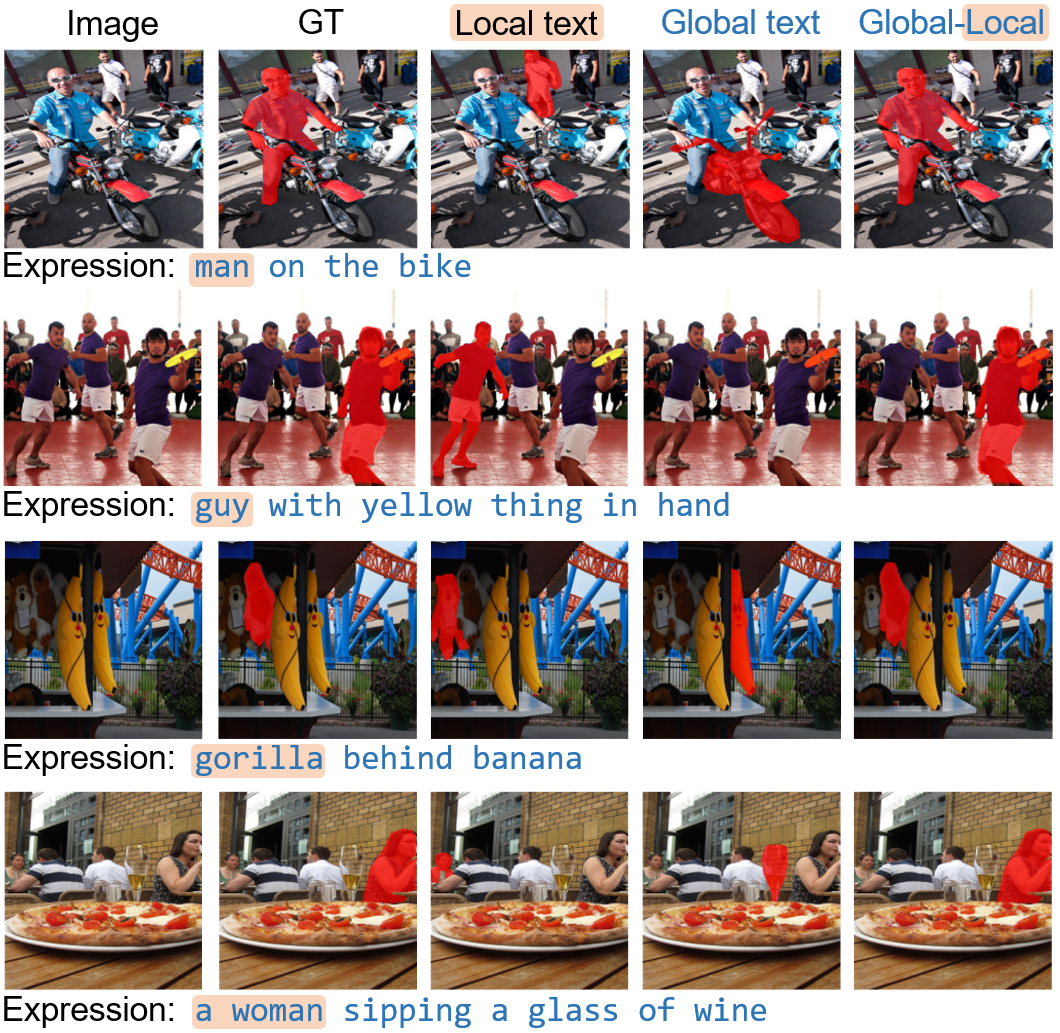}
        \caption{Additional qualitative results with different levels of textual features using COCO Instance GT mask proposals. Our Global-Local textual feature can predict target objects better than other textual features.}
        \label{fig:suple_qual_text}
    \end{figure}

\paragraph{Quantitative supports for our qualitative results.}

We report quantitative results on RefCOCOg that support our qualitative results on the effects of global-local context features in Table~\ref{tab:quantitative_supports}.
To do this, we first compute mask-class accuracy (MC-ACC), the ratio of matching object classes between GT mask and predicted mask.
Note that each GT mask is labeled with an object class and the class of a predicted mask is determined by the GT mask with the largest IoU.
Then, we compute oIoU on a subset containing only those examples whose predicted mask class is correct; this metric is dubbed as class-conditioned oIoU (CC-oIoU).
CC-oIoU captures instance localization performance when predicted mask classes are correct. 
The results first show that the global feature shows a lower MC-ACC than the local feature. 
This supports our qualitative finding that global features often select objects of a different class because it is confused by other objects present in the global context.
On the other hand, the local feature achieves lower CC-oIoU indicating more incorrect instances due to the lack of global context.
In contrast, our global-local feature allows to take advantage of both features and therefore it achieves high MC-ACC and CC-oIoU leading to a significant improvement in the final oIoU.

\begin{table}[hbt!]
\caption{Mask-class accuracy and class-conditioned oIoU with different feature types on RefCOCOg.}
\centering
\begin{tabular}{l|ccc}
\hline
Feature type & {MC-ACC}   & {CC-oIoU} & {oIoU}  \\ \hline
Global                                                                    & 78.90 & 33.10    & 27.03 \\
Local                                                                     & 83.36 & 29.94    & 25.23 \\
Global-Local                                                              & \textbf{84.32} & \textbf{35.61}    & \textbf{31.11} \\ \hline
\end{tabular}
\label{tab:quantitative_supports}
\end{table}

    \begin{figure*}
        \centering
        \includegraphics[width=1\linewidth]{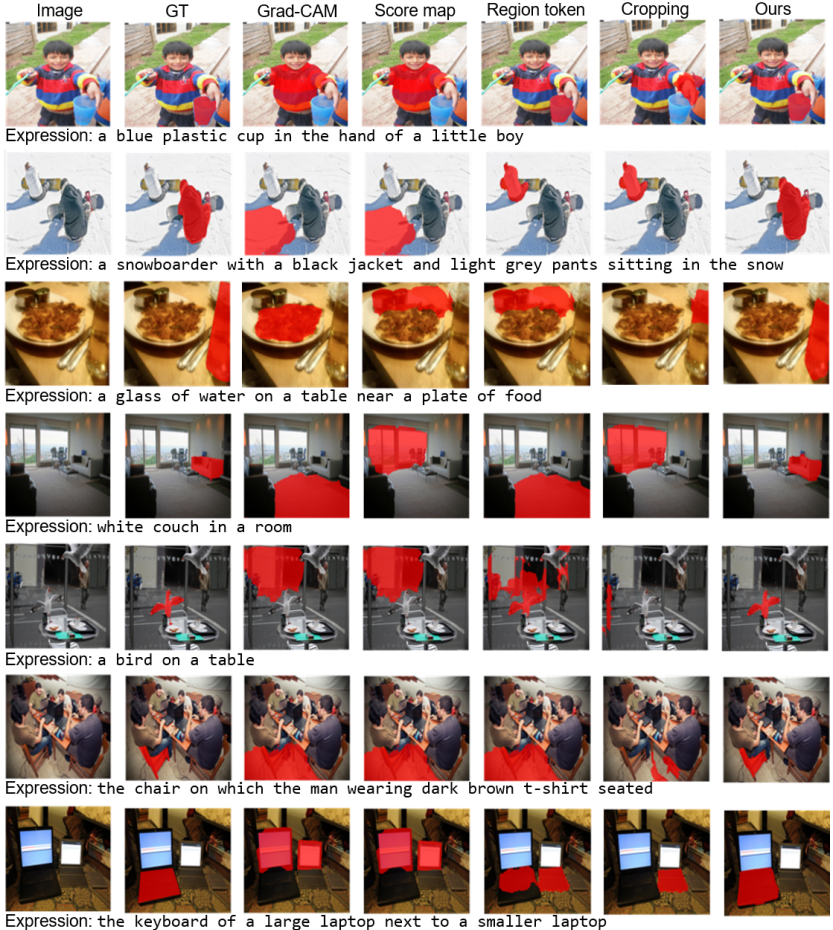}
        \caption{Qualitative results of our method with the several baselines. Note that all methods use mask proposals generated by FreeSOLO.}
        \label{fig:more_qualitative}
    \end{figure*}








\fi


\end{document}